\documentclass{article}
\usepackage[english]{babel}

\usepackage[a4paper,top=2cm,bottom=2cm,left=3cm,right=3cm,marginparwidth=1.75cm]{geometry}

\usepackage{amsfonts}
\usepackage{amsmath, bm}
\usepackage{graphicx}
\usepackage{algorithm}
\usepackage[mathscr]{euscript}
\usepackage[noend]{algpseudocode}
\usepackage[colorlinks=true, allcolors=blue]{hyperref}
\usepackage{verbatim} 
\usepackage{multirow}
\usepackage{booktabs}

\title{No GPU? No problem: an ultra fast 3D detection of road users with a simple proposal generator and energy-based out-of-distribution PointNets}
\author{Alvari Seppänen, Eerik Alamikkotervo, Risto Ojala, Giacomo Dario, Kari Tammi}

\begin{document}
\maketitle

\begin{abstract}
\noindent This paper presents a novel architecture for point cloud road user detection, which is based on a classical point cloud proposal generator approach, that utilizes simple geometrical rules.
New methods are coupled with this technique to achieve extremely small computational requirement, and mAP that is comparable to the state-of-the-art. 
The idea is to specifically exploit geometrical rules in hopes of faster performance. 
The typical downsides of this approach, \textit{e.g.} global context loss, are tackled in this paper, and solutions are presented.
This approach allows real-time performance on a single core CPU, which is not the case with end-to-end solutions presented in the state-of-the-art. 
We have evaluated the performance of the method with the public KITTI dataset, and with our own annotated dataset collected with a small mobile robot platform. 
Moreover, we also present a novel ground segmentation method, which is evaluated with the public SemanticKITTI dataset. 
\end{abstract}

\section{Introduction}

Road user detection is an important task for an autonomous mobile robot, and it plays a key role \textit{e.g.} in obstacle avoidance and route planning tasks. 
Sensors that produce point clouds are useful due to the rich 3D information that they provide. 
Therefore, they are common in mobile robot applications. 
These sensors are \textit{e.g.} LiDAR and depth camera, or even monocular camera, thanks to impressive advancements in image processing methods. 
As road user detection is a safety critical task, low latency and high mAP are desired. 
Earnest efforts have been directed towards the development of methods that achieve high mAP.
However, significantly less effort has been directed towards low computational approaches.
Example benefits of an algorithm that has low computational requirement in mobile robot applications are:

\begin{itemize}
    \item less energy consumption results in longer operation time,
    \item higher FPS increases safety,
    \item no need for an expensive processing unit,
    \item a GPU is not a requirement, which results in lighter weight, which is crucial especially for aerial robots, 
    \item more processing resources can be allocated for other functions of the robot.
\end{itemize}

\begin{figure}[H]
\centering
\includegraphics[width=0.5\textwidth]{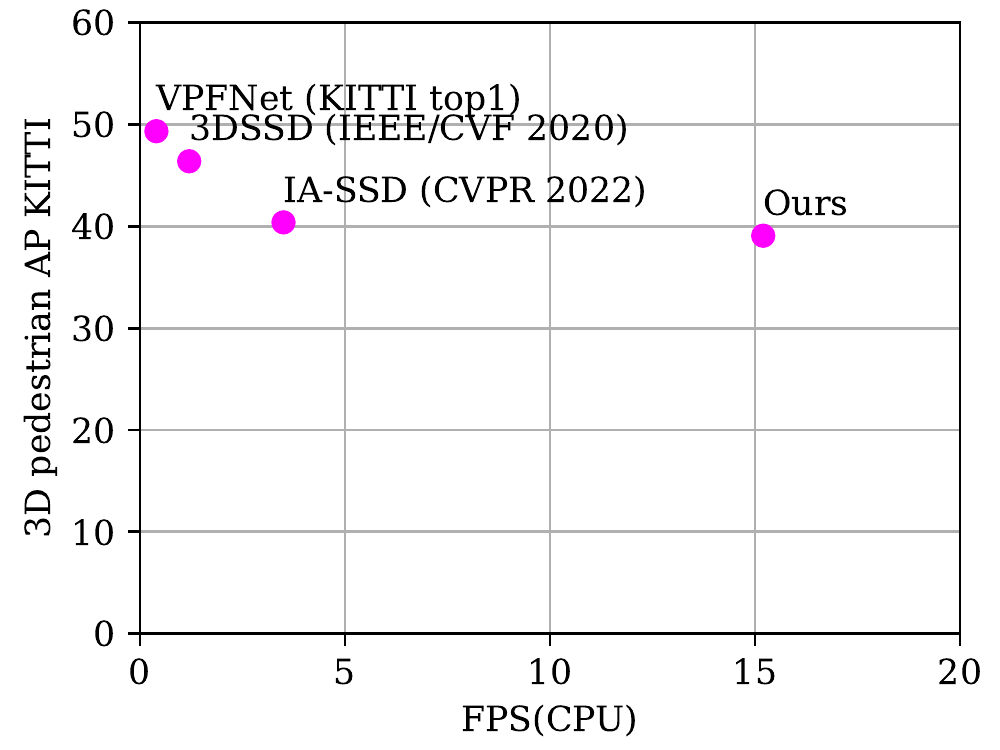}
\caption{\label{ap_vs_fps} Our method compared to the state-of-the-art.}
\end{figure}

This paper takes a drastically different approach for 3D road user detection to achieve low computational requirement. 
Unlike most method is the literature, our approach discards information, while increasing algorithmic complexity.
In other words, the first pieces of the algorithm are simple to quickly discard substantial amounts of irrelevant data.
As the amount of data decreases, the complexity of the algorithm increases. 
We implemented this idea by utilizing a classical proposal generator and energy-based out-of-distribution PointNets.
With this approach, we can achieve impressive results in terms of FPS, and comparable AP in 3D pedestrian detection to state-of-the-art.
Figure \ref{ap_vs_fps} shows the comparison to the state-of-the-art in terms of 3D pedestrian detection AP on KITTI dataset and FPS.


\subsection{Related work}



Point cloud object detection is a well-established field. 
The detection methods are mainly divided into voxel, range image, and bird's eye view methods. 
VoxelNet \cite{zhou2018voxelnet} partitions the point cloud into voxel partitions. 
Points within a voxel partition are encoded into a vector representation characterizing the shape information. 
3D convolution is performed on the voxels to predict bounding boxes and classes. 
SECOND \cite{yan2018second} improved the accuracy and reduced computational load of VoxelNet by preprocessing the point cloud by dropping voxels that include no points.
 
Authors of \cite{fan2021rangedet, chai2021point, liang2020rangercnn, meyer2019lasernet} use range image –based method. 
A range image is a projection of the point cloud on a spherical surface. 
The benefit of using projection is to preserve the information of neighboring measurements. 
However, the authors claim that some of the information is lost during the projection. 
The authors use 2D convolutional neural network to make predictions for bounding boxes and classes. 

Third popular approach is to use bird’s eye view pseudo image as input for the detection algorithm \cite{lang2019pointpillars, yang2018pixor, zheng2021se, zheng2020cia}. 
PointPillars \cite{lang2019pointpillars} construct pillar-like features from the point cloud with neural network and forms a bird’s eye view pseudo image. 
Then a 2D convolutional neural network is applied to this image, which makes the predictions. 
Bird’s eye view is a convenient representation of the point cloud because objects are rarely stacked on top of each other in, for example, outdoor driving scenarios, given an optimal vertical field-of-view of the sensor.

Li \textit{et al.} \cite{li2020real} focused on the proposal classification with limited processing resources. 
They presented a method where the proposals are generated by removing the points belonging to the ground and then clustering the remaining points.
The proposals are classified using a neural network. 
They achieved impressive results in terms of recall given the computational limitation, even comparable to the state-of-the-art.

\subsection{Scientific contribution}

The literature has a large amount of point cloud object detection methods that are tested for road user detection.
However, most of them require a powerful GPU unit to run in real-time, which some mobile robots do not have due to \textit{e.g.} power, cost, weight, or size restrictions. 
Therefore, we revisit a classical approach where object proposals are generated with a simple algorithm, which runs in real-time on a standard \textbf{C}PU and introduce novel methods for making the detections from the simple proposals, while preserving the light computational requirement. 
These novel methods are the contributions of this paper, and they are summarized in descending order below. 

\begin{enumerate}
    \item A novel architecture for 3D detection of road users on point cloud data, which include following subsections: 
    \item the implementation of an energy-based out-of-distribution detection and in-distribution multi-class classification neural network for point cloud object proposals in traffic scenarios,
    \item the extension of the conventional classification of proposals with a proposal voxel location encoder, which improves the accuracy of the model by a significant margin,
    \item study on PointNet critical and upper bound point sets of road users in classification and bounding box tasks and how they can be used for improving the out-of-distribution detection, and
    \item a ground segmentation method, which outperforms competitive methods in literature. 
    We present simple convolutional filters for the sampling of ground points for the plane fit.
    This simple but effective sampling method does not exist in the literature to the best of authors knowledge. 
\end{enumerate}

\subsection{Paper structure}

First, we present the methodology in detail, then the implementation details and experimental setup, then, the results on public and private datasets.
We also present some ablation studies, qualitative results, and discussion. 
Finally, we conclude the paper and propose future research directions. 

\section{Methods}

\begin{figure}[H]
\centering
\includegraphics[width=1.0\textwidth]{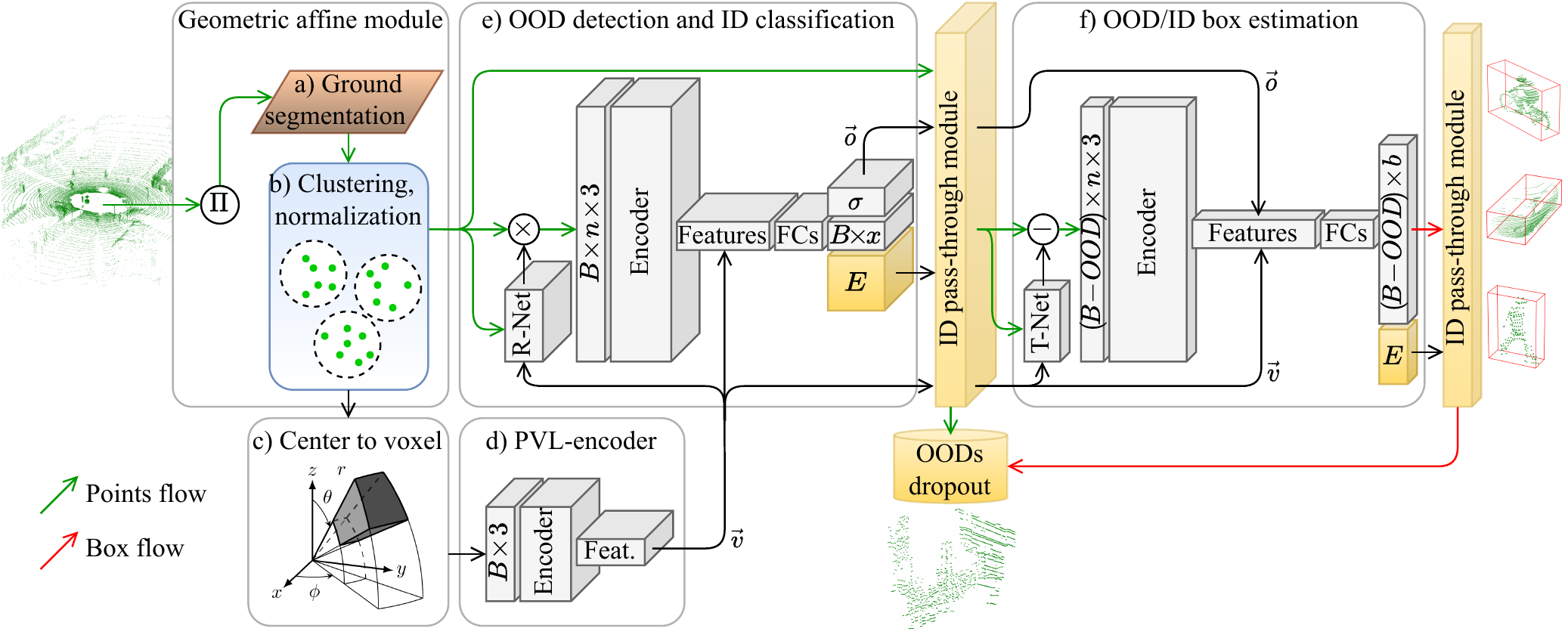}
\caption{\label{fig:pipeline} \textbf{The proposed architecture}. The input point cloud is organized by mapping $\Pi :\mathbb{R}^{n \times 3} \mapsto \mathbb{R}^{s_h \times s_w \times 3}$. Then, the ground segmentation coupled with a clustering algorithm generate simple proposals, that are fed into the neural networks. An ID pass-through module discards the proposals that do not belong to the ID samples, which enables low computational requirement for the box estimation network. Finally, a second ID pass-through module discards boxes that do not belong to the ID samples. The output of the pipeline are 3D bounding boxes and class probabilities for the objects of interest.}
\end{figure}

A general schematic of the proposed architecture is presented in Figure \ref{fig:pipeline}. 
The basic principle is to generate simple proposals and utilize effective classifiers that differentiate between the in-distribution (ID) proposals and the out-of-distribution (OOD) proposals. 
This is implemented using novel energy-based OOD PointNets.
The first PointNet predicts the class probability vector and OOD/ID energy score for the proposals, and the first batch of OODs are discarded. 
Then, 3D bounding boxes are predicted to the remaining samples with an alternative PointNet, which also predicts ID/OOD energy score for the samples.
Moreover, a novel proposal voxel location encoder (PVLE) is utilized to preserve information that would otherwise be lost in the proposal normalization process.
This encoder attempts to increase the accuracy of the neural networks.
The increased accuracy will prove that proposal location information has value in the proposal classification and bounding box estimation tasks. 
To summarize, the architecture aims to achieve low computational requirement by effective information discarding in hierarchical manner, while preserving the useful information regarding to the road user detection task.

\subsection{Ordered point cloud representation}

Points from a typical LiDAR sensor $\bm{p} = \langle x, y, z\rangle$ are mapped $\Pi :\mathbb{R}^{n \times 3} \mapsto \mathbb{R}^{s_h \times s_w \times 3}$ to spherical coordinates, and finally to image coordinates, as defined by

\begin{align}
    \begin{bmatrix}
        p_u \\
        p_v 
    \end{bmatrix}
    =
    \begin{bmatrix}
        1/2(1-\tan ^{-1}(yx^{-1})\pi ^{-1}) s_w \\
        (1- (\sin ^{-1}(z \sqrt{x^2 + y^2 +z^2}^{-1}) + f_{\mathit{up}})f^{-1}) s_h
    \end{bmatrix}
\end{align}

\noindent where $(u, v)$ are image coordinates, $(s_h, s_w)$ are the height
and width of the desired projection image representation, $f$ is the total vertical field-of-view of the sensor, and $f_{up}$ is the vertical field-of-view spanning upwards from the horizontal origin plane.
The resulted list of image coordinates is used to construct a $(x, y, z)$-channel image, which is the input for the next stage of the proposed architecture. 

\subsection{Ground segmentation}

\begin{figure}[H]
\centering
\includegraphics[width=0.5\textwidth]{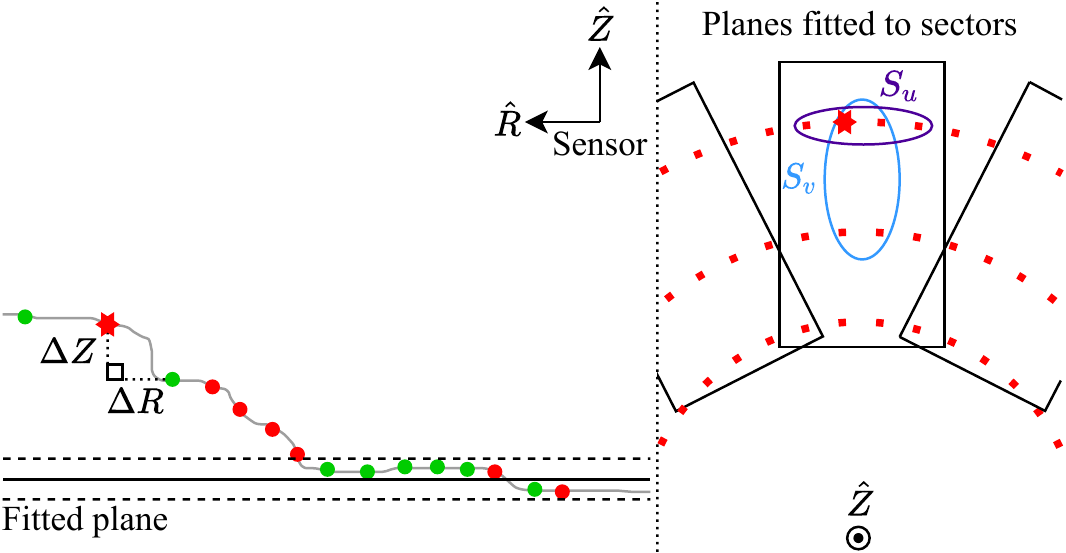}
\caption{\label{fig:ground}An illustration of the ground segmentation method. \textbf{On the left}, points that passed the filter are indicated by green, and points that are between the dashed lines are segmented ground points. \textbf{On the right}, points surrounded by ellipses are considered by $\bm{S}_v$ and $\bm{S}_u$ filters when computing the point marked with a star. A plane is fitted to each sector of points. Note that here $\bm{R} = \sqrt{\bm{X}^2 + \bm{Y}^2}$ \textit{i.e.} the distance to the $Z$-axis.}
\end{figure}

As an additional contribution, we introduce a novel ground segmentation method, which is ultra fast and more accurate than the methods existing in the literature.
Our ground segmentation combines a novel point sampling with the well proven RANSAC plane fitting method.
Figure \ref{fig:ground} shows a graphical presentation of the method.

Ground segmentation is done to differentiate between the ground and the object proposals. 
It is essential to segment the ground in the beginning of the pipeline for following reasons:
\textbf{a)} reduce amount of points to process, 
\textbf{b)} remove points that are invalid and not considered in later stages of the pipeline, and
\textbf{c)} improve the performance of the clustering algorithm, as points in some clusters are close to the points of the ground plane.

We sample the potential ground plane points with two convolutional Sobel inspired filters \cite{sobel2014}.
These kernel filters are formulated as follows

\begin{align}
    \bm{S}_v = 
    \begin{bmatrix}
    2 & 1 \\
    -2 & -1
    \end{bmatrix}
    &&
    \bm{S}_u = 
    \begin{bmatrix}
    1 & 2 & -2 & -1
    \end{bmatrix}.
\end{align}

\noindent The filters are discrete differentiation operators that operate on the point cloud projection image. 
$\bm{S}_v$ yields an approximation of vertical derivative on the projection surface, and $\bm{S}_u$ gives us a horizontal approximation.
The first term of $\bm{S}_v$, and second term of $\bm{S}_u$ are the centers of the filters, respectively.
The filters incorporate information of multiple neighboring points with Gaussian function, where points closer to center have higher effect, because simply computing the derivatives with subtraction of a single neighboring point is not viable due to noise in a typical LiDAR measurement.
In a typical LiDAR sensor, the horizontal resolution is significantly higher than the vertical resolution, therefore, the shape of $\bm{S}_u$ is 1x4 which means that it does not consider points on neighboring rows, which are significantly more distant compared to points on neighboring columns.
This way, filter is able to capture the approximation of local derivative more accurately.
LiDAR point cloud pattern on a planar ground is visualized on the right side of Figure \ref{fig:ground}.
Convolutions are run on the range matrix $\sqrt{\bm{X}^2 + \bm{Y}^2} = \bm{R} \in \mathbb{R}^{s_h\times s_w}$ and height matrix $\bm{Z} \in \mathbb{R}^{s_h\times s_w}$.

\begin{align}
    \bm{F}_y = \frac{\Delta \bm{Z}}{\Delta \bm{R}} = \frac{\bm{S}_v * \bm{Z}}{\bm{S}_v * \bm{R}},   \bm{R}(r, c) \ne 0
    &&
    \bm{F}_x = \bm{S}_u * \bm{R}
\end{align}

\noindent where $*$ denotes the 2-dimensional convolution operation.
Matricies $\bm{F}_y$ and $\bm{F}_x$ denote an approximation of point-wise normal, as filters produce derivatives.
This method for approximating point-wise normal requires only a small amount of computation, still preserving satisfactory accuracy.
We apply a threshold to $\bm{F}_y$ and $\bm{F}_x$, which gives us a sample mask $\bm{F}_{\mathit{mask}}$ of ground points, which will be the input for the RANSAC algorithm:

\begin{align}
    \bm{F}_{\mathit{mask}}(r, c) =
    \begin{cases}
        1, & \text{if } -f_{th1}\leq \bm{F}_y(r, c)\leq f_{th1}, \text{ and} -f_{th2}\leq \bm{F}_x(r, c)\leq f_{th2} \\
        0, & \text{otherwise}.
    \end{cases}
\end{align}

\noindent Then, an element-wise matrix multiplication transformation into $3 \times n$ matrix is carried out to compute the sample of ground points $\bm{G}_{\mathit{samples}} \in \mathbb{R}^{n \times 3}$ for the RANSAC algorithm:

\begin{align}
    \bm{G}_{\mathit{samples}} = \bm{F}_{\mathit{mask}} \odot \langle \bm{X}, \bm{Y}, \bm{Z}\rangle \times \bm{1}^T
\end{align}

\noindent where $\langle \bm{X}, \bm{Y}, \bm{Z}\rangle \in \mathbb{R}^{s_h \times s_w \times 3}$ contain the Cartesian coordinates of each point in the point cloud, and $\odot$ indicates Hadamard product. 
Only the sampled points are considered as the inliers for the RANSAC algorithm, thus only handful of iterations are needed, compared to case where the inliers are searched from the entire point cloud.
Again, saving a lot of computation. 
$\mathit{RANSAC}(\bm{G}_{\mathit{samples}}) = \langle a_1, a_2, a_3, a_4\rangle$ gives the parameters of the detected plane.
The distances $\bm{D} \in \mathbb{R}^{s_h\times s_w}$ between points and the plane is computed using the following equation:

\begin{align}
    \bm{D} = \frac{a_1 \bm{X} + a_2 \bm{Y} + a_3 \bm{Z} + a_4}{\sqrt{a_1^2 + a_2^2 + a_3^2}}.
\end{align}

\noindent Finally, a threshold operation defines all points belonging to the ground $\bm{G}_{\mathit{mask}} \in \mathbb{Z}_2^{s_h\times s_w}$:

\begin{align}
    \bm{G}_{\mathit{mask}}(r, c) =
    \begin{cases}
        1, & \text{if } |\bm{D}(r, c)| < p_{th} \\
        0, & \text{otherwise}.
    \end{cases}
\end{align}

\subsection{Point cloud clustering}

We are using the depth cluster method \cite{bogoslavskyi2016fast}, mainly because it is fast (full scan from a 64 channel LiDAR in 10 ms with a single core of a 2.0 GHz CPU, reported in \cite{bogoslavskyi2016fast}).
However, our architecture does not have constrains that would prevent the usage of other standard clustering algorithms such as \cite{zhao2021technical, rusu2010semantic, rusu20113d, papon2013voxel, zermas2017fast}. 
We picked the depth cluster method simply because it runs the fastest on our device.
The depth cluster algorithm computes the angle between neighboring points, and if this angle satisfies a threshold, the point in hand is assigned under the label of the currently computed cluster. 
A breath-first search (BFS) is implemented to add points to the current cluster. 
Completed BFS indicates a completion of a cluster. 
The algorithm is fast since is takes advantage of the order of the point cloud, which means that finding the neighboring points is convenient.
Unfortunately, organized point clouds have holes, which are caused by the missing reflections of the laser.
Therefore, algorithm searches points not only from instant neighbors but from a region of few points. 

\subsection{Road user detection from the simple proposals}

This paper presents a novel energy-based OOD classification and ID multi-class classification and 3D bounding box estimation networks for point clusters in traffic scenarios. 
The proposed neural network architectures build on the works of the creators of PointNet, Qi \textit{et al.} \cite{qi2017pointnet, qi2018frustum}, applying some application specific modifications in order to reduce computational requirement.
These modifications include the concatenation of cluster voxel position encoder features, to leverage on the observation angle and distance, and simplified architecture in the main encoder as well as in the fully connected layers. 
The main modification is the implementation of an energy-based OOD learning objective to mitigate the false positive rate, since our proposal generator is simple, resulting in vast amounts of OOD instances. 
Moreover, the amount of safety critical classes present in typical traffic scenarios is lower than the total amount of classes in the original PointNet \cite{qi2017pointnet}, which enables major simplification of the networks, which results in the decrease in computational complexity.
Lastly, we discovered that the respective critical and the upper bound points sets for classifier and box estimation networks are different for same cluster of points.
We take advantage of this phenomenon by implementing two separate ID pass-through modules to the pipeline for more effective OOD/ID separation. 



\subsubsection{Multi-task learning with an energy-based loss function}

The basic idea of an energy-based function is to map each point of the input space to a non-probabilistic scalar called energy $E(\bm{x}; f):\mathbb{R}^D \mapsto \mathbb{R}$ \cite{lecun2006tutorial}. 
In our application, the output vectors of the classifier and bounding box estimator networks are mapped into their respective energy scalars, that represent the "in-distributioness" of a cluster of points for a given task.
Method presented here is based on Liu \textit{et al.} \cite{liu2020energy}, where modified version of the \textit{Helmholtz free energy} from statistical mechanics is used as $E(\bm{x}; f)$ \cite{hinton1993autoencoders}.


\begin{align}
    E(\bm{x};f) = -T \cdot \log \sum ^K_i e^{f_i(\bm{x})/T} \label{energy_eq}
\end{align}

\noindent where $\bm{x}$ denotes the logits of a neural network, $T$ temperature scalar, $K$ number of logits, and $f$ a neural network.
The energy-based training objective for the classifier is same as in \cite{liu2020energy}.

\begin{align}
    \min _\theta \; \mathbb{E}_{(\bm{x}, y) \sim \mathscr{D}^{\mathit{train}}_{\mathit{ID}}}(-\log \cdot F_y(\bm{x}))+\lambda \cdot \mathscr{L}_{\mathit{energy}}
\end{align}

\noindent where energy loss is computed as

\begin{align} \label{energy_loss}
    \mathscr{L}_{\mathit{energy}} = \mathbb{E}_{(\bm{x}_{\mathit{ID}}, y) \sim \mathscr{D}^{\mathit{train}}_{\mathit{ID}}}((E(\bm{x}_{\mathit{ID}}; f) - m_{\mathit{ID}})^+)^2 \nonumber \\
    + \mathbb{E}_{\bm{x}_{OoD} \sim \mathscr{D}^{\mathit{train}}_{\mathit{OoD}}}((m_{\mathit{OoD}} - E(\bm{x}_{\mathit{OoD}}; f))^+)^2
\end{align}

\noindent where $\bm{x}_{\mathit{ID}}$ is an ID sample from KITTI training split (point inside a ground truth bounding box), and $\bm{x}_{\mathit{OOD}}$ is an OOD sample from the auxiliary OOD dataset. 
$m_{\mathit{ID}}$ and $m_{\mathit{OOD}}$ are the means of ID and OOD energies, respectively. 
The training objective is different for the box estimation network.
Loss function for the box estimation network

\begin{align}
    \mathscr{L}_{\mathit{box}} = \mathscr{L}_{\mathit{c1-reg}} + \mathscr{L}_{\mathit{c2-reg}} + \mathscr{L}_{\mathit{h-cls}} + \nonumber \mathscr{L}_{\mathit{h-reg}} + \\ \mathscr{L}_{\mathit{s-cls}} + \mathscr{L}_{\mathit{s-reg}} + \gamma \mathscr{L}_{\mathit{corner}} + \lambda \mathscr{L}_{\mathit{box-energy}}
\end{align}

\noindent where corner loss is computed as

\begin{align}
    \mathscr{L}_{\mathit{corner}} = \sum _{i=1}^{\mathit{NS}} \sum _{j=1}^{\mathit{NH}} \delta _{ij} \min \left\{ \sum _{k=1}^{8} ||P_{k}^{ij} - P_{k}^{*}||, \sum _{i=1}^{8} ||P_{k}^{ij} - P_{k}^{**}||\right\} 
\end{align}

\noindent box estimator energy $\mathscr{L}_{\mathit{box-energy}}$ is computed with the Equation (\ref{energy_loss}), where logits are defined as the heading and size class logits from the output vector $f_b(\bm{x})$.

\subsubsection{ID pass-through modules}

During inference time, energy score for each sample is computed from logits of the PointNets using equation (\ref{energy_eq}).
This is straightforward with the classifier PointNet, since logits $f_c(\bm{x})$ are just the class probabilities.
However, the output of the box estimation network $f_b(\bm{x})$ is a structure of heading and size class probabilities, and residual heading, size, and center predictions. 
Consequently, we have to find optimal way of using this special vector.
We found experimentally that the heading and the size class provide the best measure for ID/OOD separation.
We also discovered that the critical and upper bound point sets for an input $\bm{x}$ are significantly different in the classification and box estimation tasks, respectively.
Therefore, we implemented two separate ID pass-thorough modules, to have more effective ID/OOD separation.

\begin{figure}[H]
\centering
\includegraphics[width=.48\textwidth]{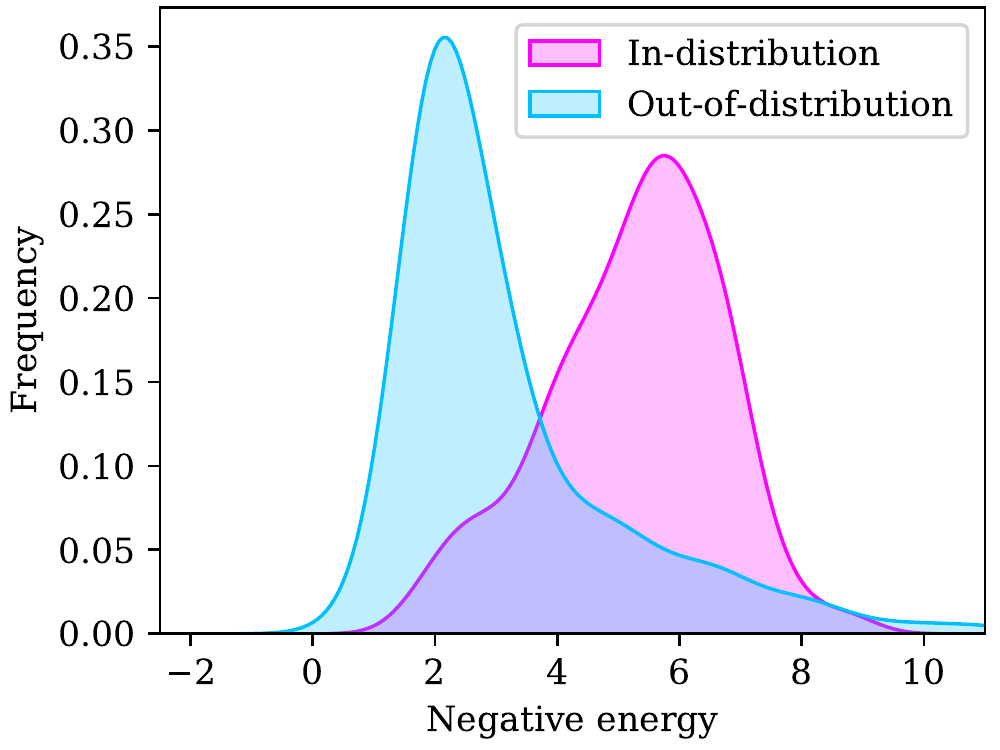}
\includegraphics[width=.48\textwidth]{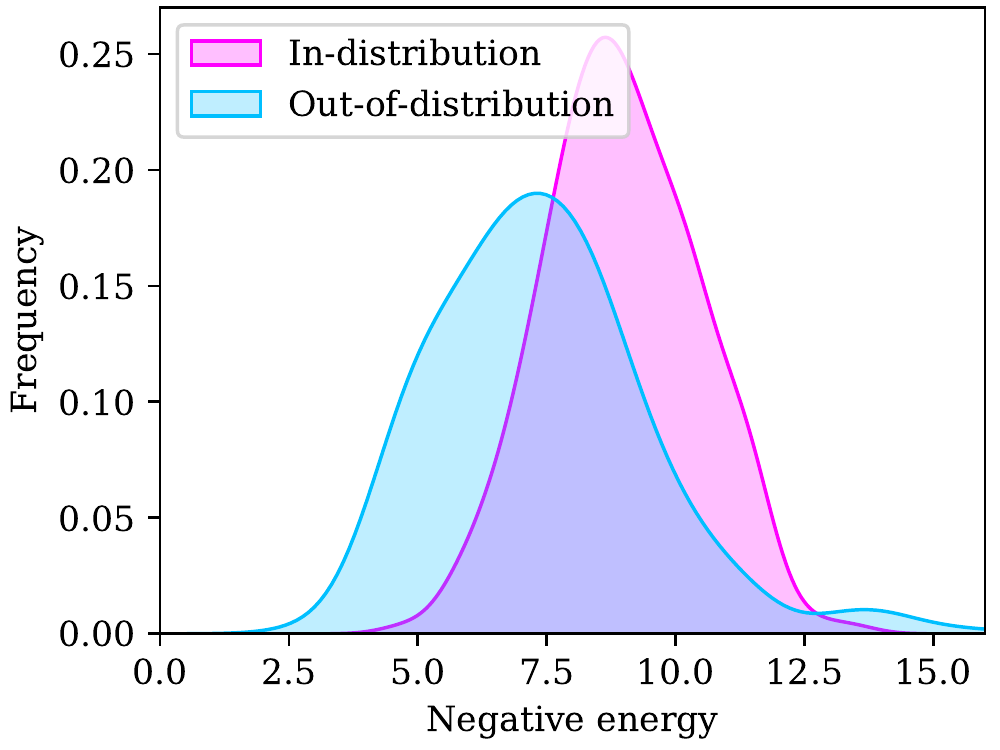}
\caption{The heading (right) and size (left) class vectors for a pre-trained box estimation network.}
\label{fig:heading_size_pre_trained}
\end{figure}

The ID/OOD classification is computed with a threshold $\gamma$, which is a chosen to classify 95\% of the ID samples correctly.

\begin{align}
    p(\bm{x}; \gamma _c, \gamma _b, f_c, f_b) = 
    \begin{cases}
        \text{in}, & \text{if } E_c(\bm{x}; f_c) < \gamma _c, \text{ and } E_b(\bm{x}; f_b) < \gamma _b\\
        \text{out}, & \text{otherwise }\!\! .
    \end{cases}
\end{align}

\subsection{Proposal voxel location encoding}

Proposals, \textit{i.e.} point clusters, are normalized before the classifier because the performance is better with canonical inputs \cite{qi2017pointnet}. 
However, location information (observation angle and distance) is lost during this process.
To make use of this information, we propose a proposal voxel location encoder (PVLE) module, which aims to improve the point cloud proposal classification and 3D bounding box estimation tasks.
The module processes proposal mean point voxel coordinates and outputs learned features.

The intuition behind this method is that the observation angle and the distance of an ID proposal carry useful information, since proposal shape, in our application, is never dominated by a concave feature, and the density is some function of down sampling and $1/r_m^2=1/(x_m^2+y_m^2+z_m^2)$, where subscript $m$ denotes a mean point of a proposal, and $r$ denotes the Euclidean distance to the origin.
The coordinates of the mean point of given proposal cluster is first voxelized and then encoded into a small feature vector, which is concatenated with the global features of the classifier and bounding box networks as well as the R-Net and T-Net (small spatial transformer networks). 
This is done so that the networks can leverage on the observation angle and distance of a given proposal.
The design of this encoder is inspired by the first encoder layer of the PointNet (MLP: 3-64).
We use a standard approach of utilizing a fully connected layer, which produces the output vector of the encoder (FC: 64-32). 
We utilize similar approach to the vanilla PointNet because voxel coordinate input has the same shape as a point input in the vanilla PointNet.

The proposal mean point is voxelized in spherical coordinates. 
We chose this system because it is the natural system for sensors, \textit{e.g.} LiDAR, which emits lasers from a single point.
Note that the point density follows the inverse-square law $d \propto 1/r^2$, voxel volume in spherical coordinate system compensates this density variation, since voxel volume follows square law $v \propto r^2$.
This results in similar magnitude of points in each voxel.
Another reason for this coordinate system is that given an uneven ground and occasional sensor pivot, proposals shift to different voxel coordinates, by utilizing spherical coordinates, it is more robust in this situation. 


\section{Experiments}

Experiments are conducted in three datasets. 
KITTI \cite{geiger2012we} and SemanticKITTI \cite{behley2021ijrr} are used to measure the accuracy of the 3D object detection and ground segmentation, respectively. 
The detection algorithm is trained on KITTI training set. 
Moreover, detection accuracy is also measured on the DBot dataset, which is our own annotated data, collected with a compact mobile robot platform, which is equiped with a Velodyne VLP-16 LiDAR. 
In-depth analysis is carried out to validate our design choices. 
Lastly, qualitative results and discussion of the strengths and weaknesses of our methods is presented in section \ref{Q_R_and_D}.




\subsection{Implementation details}

Inference time is benchmarked with a 4.0 GHz CPU.
Training is done with a NVidia GTX 1060 GPU.
We utilized following languages and libraries: PyTorch, Python, Cython, pyBind11, C++, NumPy, SciPy.
Both classifier and box estimator are trained for 200 epochs with a learning rate of 0.001, betas are 0.9 and 0.999, and we use the Adam optimizer \cite{kingma2014adam}.
The resolution of the voxel grid is: azimuth $10^{\circ}$, elevation $10^{\circ}$, and range $1$ m.

\subsection{Overall performance}

For the comparison, we chose the best method for 3D pedestrian detection of the official KITTI benchmark: VPFNet \cite{zhu2021vpfnet}, the best for 3D car detection: SFD \cite{wu2022sparse}, and state-of-the-art high FPS methods: IA-SSD \cite{zhang2022not} and 3DSSD \cite{yang20203dssd}.
The preformance on KITTI \textit{val} split is presented in Table \ref{kitti_val}.

\begin{table}[H]
\centering
\caption{3D detection AP on KITTI \textit{val} set.}
\label{kitti_val}
\begin{tabular}{ c | c c c | c c c | c c c }
 \toprule
 \multirow{2}{3em}{Method} & \multicolumn{3}{c|}{\textbf{Cars}} & \multicolumn{3}{c|}{\textbf{Pedestrians}} & \multicolumn{3}{c}{\textbf{Cyclists}} \\
 & Easy & Moderate & Hard & Easy & Moderate & Hard & Easy & Moderate & Hard \\ 
 \midrule
 VPFNet\cite{wang2021vpfnet} & 88.51 & 80.97 & 76.74 & 54.65 & 48.36 & 44.98 & 77.64 & 64.10 & 58.00 \\
 SFD \cite{wu2022sparse} & 91.73 & 84.76 & 77.92 & - & - & - & - & - & - \\
 IA-SSD \cite{zhang2022not} & 88.34 & 80.13 & 75.04 & 46.51 & 39.03 & 35.60 & 78.35 & 61.94 & 55.70 \\
 3DSSD \cite{yang20203dssd} & 88.36 & 79.57 & 74.55 & 54.64 & 44.27 & 40.23 & 82.48 & 64.10 & 56.90 \\
 \midrule
 Ours & 49.8 & 51.2 & 47.9 & 43.5 & 37.1 & 36.6 & 62.8 & 47.1 & 47.2 \\
\end{tabular}
\end{table}

\noindent Our method achieves similar AP on pedestrian and cyclist detection to other methods, which is impressive taking into account the simplicity of our approach. 
Table \ref{table_map_vs_fps} compares the performance in terms of FPS and mAP.
Our method is the only one that achieves real-time performance on a CPU, and on an edge computing device.

\begin{table}[H]
\centering
\caption{KITTI: mAP vs FPS.}
\label{table_map_vs_fps}
\begin{tabular}{ c | c c c c }
 \toprule
 Method & mAP & FPS (GPU) & FPS(Jetson Nano) & FPS (CPU) \\ 
 \midrule
 VPFNet \cite{wang2021vpfnet} & 65.99 & 10.0 & 4.2 & 0.4 \\
 SFD \cite{wu2022sparse} & 84.80 (Cars only) & 10.2 & 5.1 & 1.1 \\
 IA-SSD \cite{zhang2022not} & 62.29 & 83.0 & 8.5 & 3.5 \\
 3DSSD \cite{yang20203dssd} & 65.01 & 26.3 & 5.3 & 1.2 \\
 \midrule
 Ours & 47.02 & \textbf{120.1} & \textbf{90.0} & \textbf{15.2} \\
\end{tabular}
\end{table}

\begin{figure}[H]
\centering
\includegraphics[width=0.5\textwidth]{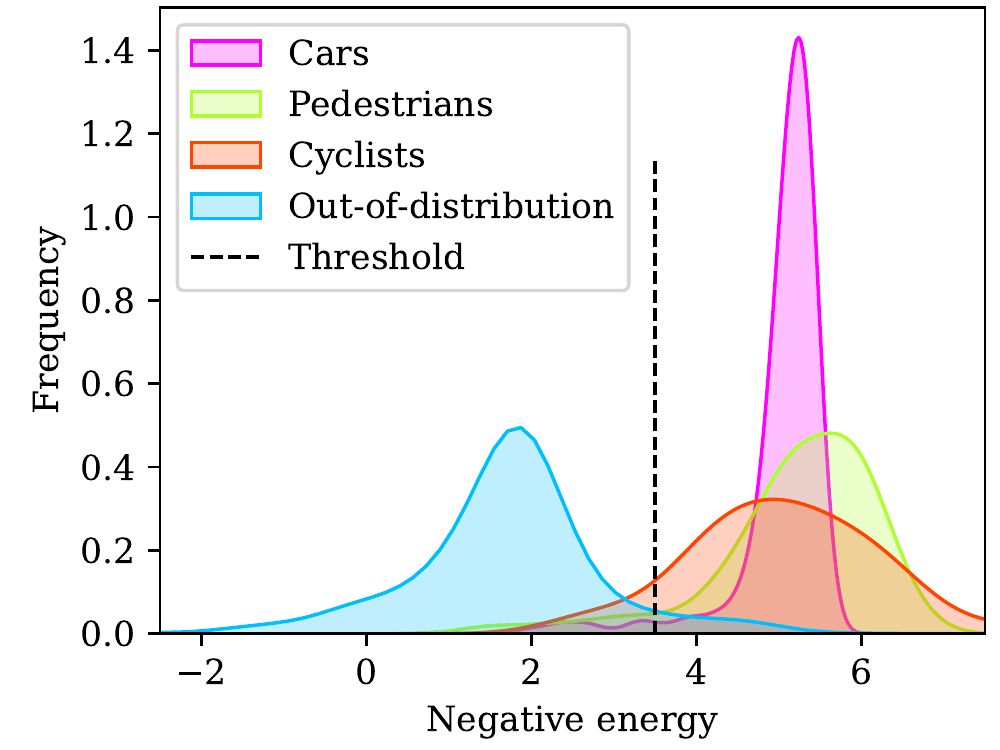}
\caption{\label{fig:class_hist} Histogram of 1000 ID and 1000 OOD samples showcases separability.}
\end{figure}

\noindent Figure \ref{fig:class_hist} illustrates the separability between ID and OOD samples.
Plot includes 1000 samples from IDs and OODs, respectively.
Note that true partitions of ID and OOD are approximately 6\% and 94\%, respectively, in the validation split.
Energy distribution of cars is much narrower compered to other classes because the amount of car samples in the training set is significantly larger compared to other classes. 
This allows the network to learn the difference between cars and OODs better.

\subsection{Ablation study}\label{ablation_study}

Table \ref{ablation_study} illustrates the contributions of the modules to the mAP and FPS(CPU).
Modules improve the mAP significantly while decrease in FPS in negligible.

\begin{table}[H]
\centering
\caption{Ablations of different modules and their contribution to the mAP and the FPS. Abbreviations: PVLE \textendash proposal voxel location encoder, IDPTM \textendash ID pass-through module}
\begin{tabular}{ c c c c | c c }
 \toprule
 PVLE(cls) & PVLE(box) & $1^{st}$ IDPTM & $2^{nd}$ IDPTM & mAP & FPS (CPU) \\ 
 \midrule
 - & - & - & - & 38.54 & 16.2 \\ 
 \checkmark & - & - & - & 39.88 & 15.6 \\
 \checkmark & \checkmark & - & - & 42.34 & 15.5 \\
 - & - & \checkmark & - & 41.32 & 15.6 \\
 - & - & \checkmark & \checkmark & 43.05 & 15.4 \\
 \checkmark & \checkmark & \checkmark & - & 45.13 & 15.3 \\
 \checkmark & \checkmark & \checkmark & \checkmark & 47.02 & 15.2 \\ 
\end{tabular}
\end{table}

\subsection{Ground segmentation}

Results of a performance tests are summarized on Table \ref{ground_results}. 
Our ground segmentation method is compared to frequent and state-of-the-art methods in the literature, and it is clearly performing better in terms of computational requirement, accuracy, and IoU.
This is due to our effective sampling method, which reduces the iterations needed in the RANSAC function.

\begin{table}[H]
\centering
\caption{Comparison of ground segmentation methods. Our results are achieved with a single core of a 4 GHz CPU, and 32 even sectors on a 360$^{\circ}$ scan, * is for GPU, - not reported, bold is best.}
\label{ground_results}
\begin{tabular}{ c | c c c c c c c }
 \toprule
 Method & Dataset & Scans & Time (s) & Precision & Recall & Accuracy & IoU \\ 
 \midrule
 HD \cite{ouyang2021pv} & semKITTI & 23201 & 0.306 & 0.47 & 0.95 & - & 0.45 \\ 
 LF \cite{ouyang2021pv} & semKITTI & 23201 & 0.658 & 0.38 & 0.77 & - & 0.34 \\ 
 GPF \cite{ouyang2021pv} & semKITTI & 23201 & 31.713 & 0.67 & 0.63 & - & 0.45 \\ 
 GPF-Opti \cite{ouyang2021pv} & semKITTI & 23201 & 0.207 & 0.66 & 0.59 & - & 0.43 \\ 
 GPF-RANSAC \cite{ouyang2021pv} & semKITTI & 23201 & 0.028 & 0.65 & 0.88 & - & 0.74 \\ 
 Hybrid-reg \cite{liu2019ground} & KITTI & 5 & 0.888 & - & - & 0.88 & - \\ 
 CNN-method \cite{velas2018cnn} & Custom & 252 & 0.139 & 0.93 & \textbf{0.99} & - & - \\
 CRF-method \cite{rummelhard2017ground}  & semKITTI & 3040 & 0,147 & 0.80 & - & - & 0.78 \\
 GndNet \cite{paigwar2020gndnet} & semKITTI & 3040 & 0.0180* & 0.84 & 0.99 & - & 0.84 \\
 \midrule
 Ours & semKITTI & 23201 & \textbf{0.00025} & \textbf{0.93} & 0.97 & \textbf{0.95} & \textbf{0.87} \\ 
\end{tabular}
\end{table}

\subsection{Qualitative results and discussion}\label{Q_R_and_D}

\begin{figure}[H]
\centering
\includegraphics[width=.32\textwidth]{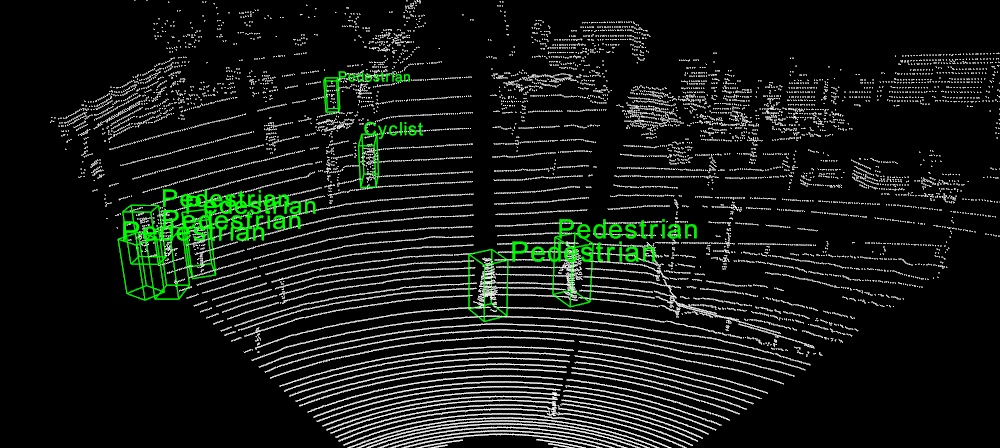}
\includegraphics[width=.32\textwidth]{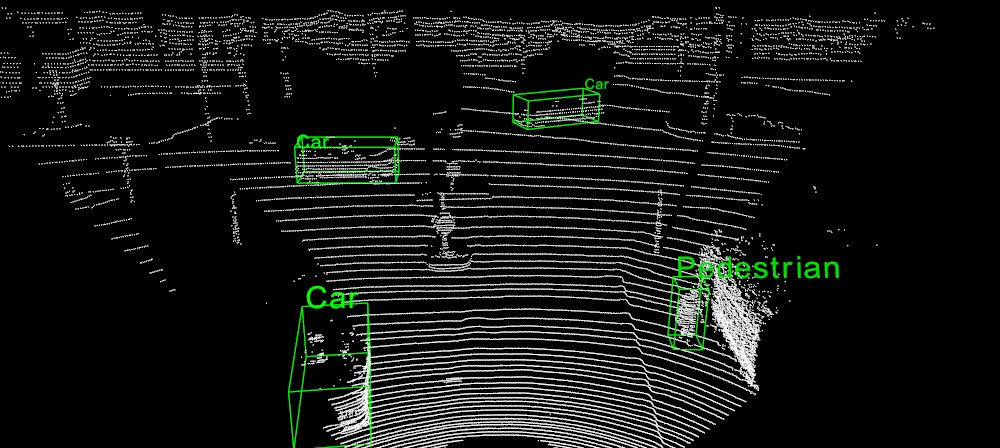}
\includegraphics[width=.32\textwidth]{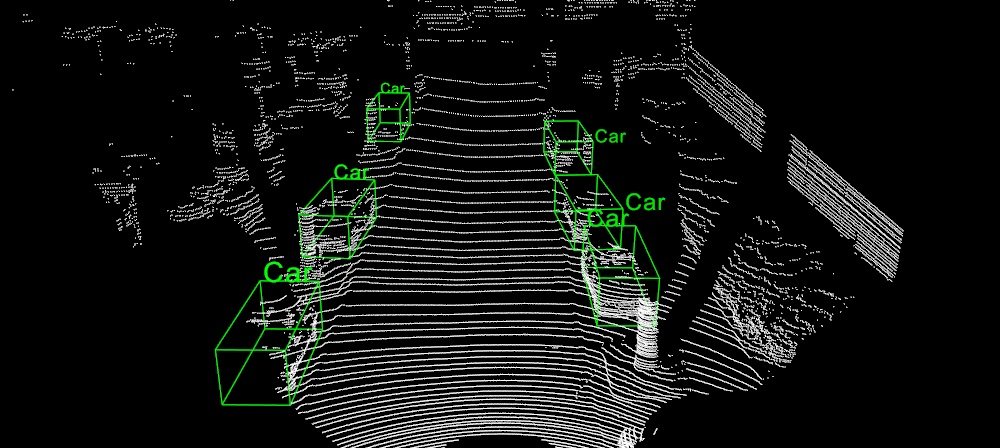}
\caption{Example detections (zoom for better detail).}
\label{fig:detection_pics}
\end{figure}

Simple proposal generator reduces the computational requirement significantly, while still achieving good mAP.
Simple proposal generator has another benefit too.
It allows data streaming \textit{i.e.} data can be processed in small sectors, which means that processing can be started earlier compared to full scan approaches.
This will decrease the latency of the detection.
The limitation of a simple proposal generator is cluster fusion when objects are close to each other.
PointNets work well as in road user classification and bounding box estimation tasks.
They can be leveraged by the proposed PVLE and energy-based OOD modules, without adding significant amount of computation.
Some of the detections are visualized on Figure \ref{fig:detection_pics}.

\section{Conclusions}

This paper presented a novel architecture for 3D road user detection task.
An impressive 15.2 FPS was achieved with a CPU only implementation while having 47 mAP on KITTI dataset.
The architecture is based on a simple proposal generator and OOD PointNets, which leverage from a novel PVLE-module.
In the future, other OOD detection methods could be studied in the 3D road user detection task.

\appendix

\bibliographystyle{unsrt}
\bibliography{bibliography}

\begin{thebibliography}{10}

\bibitem{zhou2018voxelnet}
Yin Zhou and Oncel Tuzel.
\newblock Voxelnet: End-to-end learning for point cloud based 3d object
  detection.
\newblock In {\em Proceedings of the IEEE conference on computer vision and
  pattern recognition}, pages 4490--4499, 2018.

\bibitem{yan2018second}
Yan Yan, Yuxing Mao, and Bo~Li.
\newblock Second: Sparsely embedded convolutional detection.
\newblock {\em Sensors}, 18(10):3337, 2018.

\bibitem{fan2021rangedet}
Lue Fan, Xuan Xiong, Feng Wang, Naiyan Wang, and Zhaoxiang Zhang.
\newblock Rangedet: In defense of range view for lidar-based 3d object
  detection.
\newblock In {\em Proceedings of the IEEE/CVF International Conference on
  Computer Vision}, pages 2918--2927, 2021.

\bibitem{chai2021point}
Yuning Chai, Pei Sun, Jiquan Ngiam, Weiyue Wang, Benjamin Caine, Vijay
  Vasudevan, Xiao Zhang, and Dragomir Anguelov.
\newblock To the point: Efficient 3d object detection in the range image with
  graph convolution kernels.
\newblock In {\em Proceedings of the IEEE/CVF Conference on Computer Vision and
  Pattern Recognition}, pages 16000--16009, 2021.

\bibitem{liang2020rangercnn}
Zhidong Liang, Ming Zhang, Zehan Zhang, Xian Zhao, and Shiliang Pu.
\newblock Rangercnn: Towards fast and accurate 3d object detection with range
  image representation.
\newblock {\em arXiv preprint arXiv:2009.00206}, 2020.

\bibitem{meyer2019lasernet}
Gregory~P Meyer, Ankit Laddha, Eric Kee, Carlos Vallespi-Gonzalez, and Carl~K
  Wellington.
\newblock Lasernet: An efficient probabilistic 3d object detector for
  autonomous driving.
\newblock In {\em Proceedings of the IEEE/CVF conference on computer vision and
  pattern recognition}, pages 12677--12686, 2019.

\bibitem{lang2019pointpillars}
Alex~H Lang, Sourabh Vora, Holger Caesar, Lubing Zhou, Jiong Yang, and Oscar
  Beijbom.
\newblock Pointpillars: Fast encoders for object detection from point clouds.
\newblock In {\em Proceedings of the IEEE/CVF Conference on Computer Vision and
  Pattern Recognition}, pages 12697--12705, 2019.

\bibitem{yang2018pixor}
Bin Yang, Wenjie Luo, and Raquel Urtasun.
\newblock Pixor: Real-time 3d object detection from point clouds.
\newblock In {\em Proceedings of the IEEE conference on Computer Vision and
  Pattern Recognition}, pages 7652--7660, 2018.

\bibitem{zheng2021se}
Wu~Zheng, Weiliang Tang, Li~Jiang, and Chi-Wing Fu.
\newblock Se-ssd: Self-ensembling single-stage object detector from point
  cloud.
\newblock In {\em Proceedings of the IEEE/CVF Conference on Computer Vision and
  Pattern Recognition}, pages 14494--14503, 2021.

\bibitem{zheng2020cia}
Wu~Zheng, Weiliang Tang, Sijin Chen, Li~Jiang, and Chi-Wing Fu.
\newblock Cia-ssd: Confident iou-aware single-stage object detector from point
  cloud.
\newblock {\em arXiv preprint arXiv:2012.03015}, 2020.

\bibitem{li2020real}
Xuesong Li, Jose Guivant, and Subhan Khan.
\newblock Real-time 3d object proposal generation and classification using
  limited processing resources.
\newblock {\em Robotics and Autonomous Systems}, 130:103557, 2020.

\bibitem{sobel2014}
Irwin Sobel.
\newblock An isotropic 3x3 image gradient operator.
\newblock {\em Presentation at Stanford A.I. Project 1968}, 02 2014.

\bibitem{bogoslavskyi2016fast}
Igor Bogoslavskyi and Cyrill Stachniss.
\newblock Fast range image-based segmentation of sparse 3d laser scans for
  online operation.
\newblock In {\em 2016 IEEE/RSJ International Conference on Intelligent Robots
  and Systems (IROS)}, pages 163--169. IEEE, 2016.

\bibitem{zhao2021technical}
Yiming Zhao, Xiao Zhang, and Xinming Huang.
\newblock A technical survey and evaluation of traditional point cloud
  clustering methods for lidar panoptic segmentation.
\newblock In {\em Proceedings of the IEEE/CVF International Conference on
  Computer Vision}, pages 2464--2473, 2021.

\bibitem{rusu2010semantic}
Radu~Bogdan Rusu.
\newblock Semantic 3d object maps for everyday manipulation in human living
  environments.
\newblock {\em KI-K{\"u}nstliche Intelligenz}, 24(4):345--348, 2010.

\bibitem{rusu20113d}
Radu~Bogdan Rusu and Steve Cousins.
\newblock 3d is here: Point cloud library (pcl).
\newblock In {\em 2011 IEEE international conference on robotics and
  automation}, pages 1--4. IEEE, 2011.

\bibitem{papon2013voxel}
Jeremie Papon, Alexey Abramov, Markus Schoeler, and Florentin Worgotter.
\newblock Voxel cloud connectivity segmentation-supervoxels for point clouds.
\newblock In {\em Proceedings of the IEEE conference on computer vision and
  pattern recognition}, pages 2027--2034, 2013.

\bibitem{zermas2017fast}
Dimitris Zermas, Izzat Izzat, and Nikolaos Papanikolopoulos.
\newblock Fast segmentation of 3d point clouds: A paradigm on lidar data for
  autonomous vehicle applications.
\newblock In {\em 2017 IEEE International Conference on Robotics and Automation
  (ICRA)}, pages 5067--5073. IEEE, 2017.

\bibitem{qi2017pointnet}
Charles~R Qi, Hao Su, Kaichun Mo, and Leonidas~J Guibas.
\newblock Pointnet: Deep learning on point sets for 3d classification and
  segmentation.
\newblock In {\em Proceedings of the IEEE conference on computer vision and
  pattern recognition}, pages 652--660, 2017.

\bibitem{qi2018frustum}
Charles~R Qi, Wei Liu, Chenxia Wu, Hao Su, and Leonidas~J Guibas.
\newblock Frustum pointnets for 3d object detection from rgb-d data.
\newblock In {\em Proceedings of the IEEE conference on computer vision and
  pattern recognition}, pages 918--927, 2018.

\bibitem{lecun2006tutorial}
Yann LeCun, Sumit Chopra, Raia Hadsell, M~Ranzato, and F~Huang.
\newblock A tutorial on energy-based learning.
\newblock {\em Predicting structured data}, 1(0), 2006.

\bibitem{liu2020energy}
Weitang Liu, Xiaoyun Wang, John Owens, and Yixuan Li.
\newblock Energy-based out-of-distribution detection.
\newblock {\em Advances in Neural Information Processing Systems},
  33:21464--21475, 2020.

\bibitem{hinton1993autoencoders}
Geoffrey~E Hinton and Richard Zemel.
\newblock Autoencoders, minimum description length and helmholtz free energy.
\newblock {\em Advances in neural information processing systems}, 6, 1993.

\bibitem{geiger2012we}
Andreas Geiger, Philip Lenz, and Raquel Urtasun.
\newblock Are we ready for autonomous driving? the kitti vision benchmark
  suite.
\newblock In {\em 2012 IEEE conference on computer vision and pattern
  recognition}, pages 3354--3361. IEEE, 2012.

\bibitem{behley2021ijrr}
J.~Behley, M.~Garbade, A.~Milioto, J.~Quenzel, S.~Behnke, J.~Gall, and
  C.~Stachniss.
\newblock {Towards 3D LiDAR-based semantic scene understanding of 3D point
  cloud sequences: The SemanticKITTI Dataset}.
\newblock {\em The International Journal on Robotics Research},
  40(8-9):959--967, 2021.

\bibitem{kingma2014adam}
Diederik~P Kingma and Jimmy Ba.
\newblock Adam: A method for stochastic optimization.
\newblock {\em arXiv preprint arXiv:1412.6980}, 2014.

\bibitem{zhu2021vpfnet}
Hanqi Zhu, Jiajun Deng, Yu~Zhang, Jianmin Ji, Qiuyu Mao, Houqiang Li, and
  Yanyong Zhang.
\newblock Vpfnet: Improving 3d object detection with virtual point based lidar
  and stereo data fusion.
\newblock {\em arXiv preprint arXiv:2111.14382}, 2021.

\bibitem{wu2022sparse}
Xiaopei Wu, Liang Peng, Honghui Yang, Liang Xie, Chenxi Huang, Chengqi Deng,
  Haifeng Liu, and Deng Cai.
\newblock Sparse fuse dense: Towards high quality 3d detection with depth
  completion.
\newblock {\em Accepted to CVPR 2022, arXiv preprint arXiv:2203.09780}, 2022.

\bibitem{zhang2022not}
Yifan Zhang, Qingyong Hu, Guoquan Xu, Yanxin Ma, Jianwei Wan, and Yulan Guo.
\newblock Not all points are equal: Learning highly efficient point-based
  detectors for 3d lidar point clouds.
\newblock {\em Accepted to CVPR 2022, arXiv preprint arXiv:2203.11139}, 2022.

\bibitem{yang20203dssd}
Zetong Yang, Yanan Sun, Shu Liu, and Jiaya Jia.
\newblock 3dssd: Point-based 3d single stage object detector.
\newblock In {\em Proceedings of the IEEE/CVF conference on computer vision and
  pattern recognition}, pages 11040--11048, 2020.

\bibitem{wang2021vpfnet}
Chia-Hung Wang, Hsueh-Wei Chen, and Li-Chen Fu.
\newblock Vpfnet: Voxel-pixel fusion network for multi-class 3d object
  detection.
\newblock {\em arXiv preprint arXiv:2111.00966}, 2021.

\bibitem{ouyang2021pv}
Zhenchao Ouyang, Xiaoyun Dong, Jiahe Cui, Jianwei Niu, and Mohsen Guizani.
\newblock Pv-enconet: Fast object detection based on colored point cloud.
\newblock {\em IEEE Transactions on Intelligent Transportation Systems}, 2021.

\bibitem{liu2019ground}
Kaiqi Liu, Wenguang Wang, Ratnasingham Tharmarasa, Jun Wang, and Yan Zuo.
\newblock Ground surface filtering of 3d point clouds based on hybrid
  regression technique.
\newblock {\em IEEE Access}, 7:23270--23284, 2019.

\bibitem{velas2018cnn}
Martin Velas, Michal Spanel, Michal Hradis, and Adam Herout.
\newblock Cnn for very fast ground segmentation in velodyne lidar data.
\newblock In {\em 2018 IEEE International Conference on Autonomous Robot
  Systems and Competitions (ICARSC)}, pages 97--103. IEEE, 2018.

\bibitem{rummelhard2017ground}
Lukas Rummelhard, Anshul Paigwar, Amaury N{\`e}gre, and Christian Laugier.
\newblock Ground estimation and point cloud segmentation using spatiotemporal
  conditional random field.
\newblock In {\em 2017 IEEE Intelligent Vehicles Symposium (IV)}, pages
  1105--1110. IEEE, 2017.

\bibitem{paigwar2020gndnet}
Anshul Paigwar, {\"O}zg{\"u}r Erkent, David Sierra-Gonzalez, and Christian
  Laugier.
\newblock Gndnet: Fast ground plane estimation and point cloud segmentation for
  autonomous vehicles.
\newblock In {\em 2020 IEEE/RSJ International Conference on Intelligent Robots
  and Systems (IROS)}, pages 2150--2156. IEEE, 2020.

\end{thebibliography}

\end{document}